\documentclass{article}

\usepackage{float}
\usepackage{arxiv}
\usepackage{amsmath}
\usepackage[utf8]{inputenc} % allow utf-8 input
\usepackage[T1]{fontenc}    % use 8-bit T1 fonts
\usepackage{hyperref}       % hyperlinks
\usepackage{url}            % simple URL typesetting
\usepackage{booktabs}       % professional-quality tables
\usepackage{amsfonts}       % blackboard math symbols
\usepackage{nicefrac}       % compact symbols for 1/2, etc.
\usepackage{siunitx}
\usepackage{microtype}      % microtypography
\usepackage{lipsum}		% Can be removed after putting your text content
\usepackage{graphicx}
\usepackage{doi}
\usepackage{subcaption}
\usepackage{enumitem}
\usepackage{titlesec}
\usepackage{siunitx}
\usepackage[numbers]{natbib}

\usepackage{enumitem}

\titlespacing*{\section}
  {0pt}   % left margin
  {15pt}   % space before section
  {2pt}   % space after section

\titlespacing*{\subsection}
  {0pt}   % left margin
  {15pt}   % space before section
  {2pt}   % space after section

\title{Physics-Informed Neural ODEs with Scale-Aware Residuals for Learning Stiff Biophysical Dynamics}

\author{
  \href{https://orcid.org/0000-0000-0000-0000}{\includegraphics[scale=0.06]{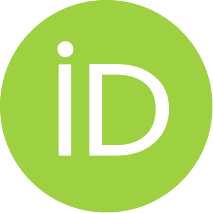}\hspace{1mm}Kamalpreet Singh Kainth} \\
  Vizuara AI Labs \\
  \texttt{-} \\
  \And
  \href{https://orcid.org/0000-0000-0000-0001}{\includegraphics[scale=0.06]{orcid.pdf}\hspace{1mm}Prathamesh Dinesh Joshi} \\
  Vizuara AI Labs \\
  \texttt{Prathmesh@vizuara.com} \\
  \And
  \href{https://orcid.org/0000-0000-0000-0001}{\includegraphics[scale=0.06]{orcid.pdf}\hspace{1mm}Raj Abhijit Dandekar} \\
  Vizuara AI Labs \\
  \texttt{raj@vizuara.com} \\
  \And
  \href{https://orcid.org/0000-0000-0000-0002}{\includegraphics[scale=0.06]{orcid.pdf}\hspace{1mm}Rajat Dandekar} \\
  Vizuara AI Labs \\
  \texttt{rajat@vizuara.com} \\
  \And
  \href{https://orcid.org/0000-0000-0000-0003}{\includegraphics[scale=0.06]{orcid.pdf}\hspace{1mm}Sreedat Panat} \\
  Vizuara AI Labs \\
  \texttt{sreedat@vizuara.com} \\
}

\renewcommand{\shorttitle}{\textit{arXiv} Template}

\begin{document}
\maketitle

\vspace{-2em}
\begin{center}
\small \textit{Accepted at the Mathematics and Scientific Machine Learning (MSML) Conference 2025.}
\end{center}

\begin{abstract}
Neural differential equations offer a powerful framework for modeling continuous-time dynamics, but forecasting stiff biophysical systems remains unreliable. Standard Neural ODEs and physics-informed variants often require orders of magnitude more iterations, and even then may converge to suboptimal solutions that fail to preserve oscillatory frequency or amplitude. We introduce Physics-Informed Neural ODEs with Scale-Aware Residuals (PI-NODE-SR), a framework that combines a low-order explicit solver (Heun’s method) with residual normalisation to balance contributions between state variables evolving on disparate timescales. This combination stabilises training under realistic iteration budgets and avoids reliance on computationally expensive implicit solvers. On the Hodgkin-Huxley equations, PI-NODE-SR learns from a single oscillation simulated with a stiff solver (Rodas5P) and extrapolates beyond \SI{100}{\milli\second}, capturing both oscillation frequency and near-correct amplitudes. Remarkably, end-to-end learning of the vector field enables PI-NODE-SR to recover morphological features such as sharp subthreshold curvature in gating variables that are typically reserved for higher-order solvers, suggesting that neural correction can offset numerical diffusion. While performance remains sensitive to initialisation, PI-NODE-SR consistently reduces long-horizon errors relative to baseline Neural-ODEs and PINNs, offering a principled route to stable and efficient learning of stiff biological dynamics.
\end{abstract}

% keywords can be removed

\section{Introduction}
Neural Ordinary Differential Equations (Neural ODEs)~\citep{DBLP:journals/corr/abs-1806-07366} have emerged as a powerful framework to learn continuous-time dynamics using neural networks. Unlike discrete-time models such as recurrent neural networks (RNNs), Neural ODEs evolve hidden states through parameterised differential equations, providing a principled and flexible way to model complex dynamical systems. This approach has influenced a wide range of domains, including time-series analysis, generative modeling, and scientific machine learning (SciML) \citep{DBLP:journals/corr/abs-2201-04630}.

Despite recent advances in neural differential equations, accurately forecasting the behavior of stiff biological systems remains a significant challenge. Stiffness arises in systems where different variables evolve on vastly different timescales, a common characteristic of many biological, chemical, and physical models. Standard Neural ODEs, while effective on smooth or moderately nonlinear dynamics, often struggle in stiff regimes where maintaining stability, efficiency, and long-term accuracy becomes critical.

This difficulty is exemplified by the Hodgkin–Huxley (HH) model~\citep{hodgkin1952}, which governs nonlinear neuronal membrane dynamics through interactions between voltage and ion channel gating variables. The HH system combines fast gating kinetics with slower recovery variables, producing oscillatory voltage dynamics that are both nonlinear and stiff. For data-driven models, this poses a dual challenge; reproducing oscillations with high short-term fidelity while maintaining stable extrapolation over long horizons.

In practice, Neural ODEs and Physics-Informed Neural ODEs (PI-NODEs) are often unreliable in such stiff regimes. Training may succeed in some runs but frequently requires orders of magnitude more iterations than practical. Even when convergence occurs, the learned dynamics often fail to capture oscillation frequency or amplitude during extrapolation, limiting their predictive utility. Physics-informed constraints help guide learning but do not always resolve the fundamental imbalance. Residuals across state variables with vastly different dynamic ranges can dominate or vanish, destabilising optimisation; gradients can become ill-conditioned, and small residual mismatches can compound over time.

To address these challenges, we develop Physics-Informed Neural ODEs with Scale-Aware Residuals (PI-NODE-SR), a framework tailored to stiff oscillatory systems. The central idea is to normalise physics residuals across variables according to their dynamic scales, ensuring that fast and slow processes contribute proportionally during training. We pair this residual normalisation with training under an explicit low-order  solver (Heun’s method). Although Heun’s method alone is typically unstable in stiff regimes, the combination with scale-aware residuals yields stable optimisation within realistic iteration budgets. This solver-loss combination allows PI-NODE-SR to capture stiff oscillatory dynamics more reliably than standard Neural ODEs or PI-NODEs.

Our contributions are as follows:
\begin{itemize}
    \item \textbf{Problem analysis} - We provide evidence that standard Neural ODEs and Physics Informed Neural Networks are unreliable for stiff oscillatory dynamics, often requiring excessive training iterations and frequently converging to suboptimal solutions.

    \item \textbf{Methodology} - We introduce scale-aware residuals, a principled normalisation strategy that stabilises physics-informed training by balancing contributions across variables with disparate timescales.

    \item \textbf{Solver-loss synergy} - We demonstrate that combining a low-order explicit solver (Heun’s method) with scale-aware residual normalisation yields stable training on stiff biophysical dynamics, despite Heun’s method alone being unsuitable for stiff systems.

    \item \textbf{Empirical validation} - On the Hodgkin–Huxley equations, PI-NODE-SR learns from a single oscillation simulated with Rodas5P and extrapolates beyond \SI{100}{\milli\second}, accurately recovering oscillation frequency and near-correct amplitudes; a regime where baselines degrade or fail.

    \item \textbf{Robustness Analysis and Ablation Study} - While sensitive to initialisation, PI-NODE-SR consistently outperforms baseline Neural ODEs and PI-NODEs across runs, highlighting an open challenge for developing initialisation strategies and training schemes for stiff, physics-informed models.
\end{itemize}

In summary, PI-NODE-SR demonstrates that scale-aware physics residuals can substantially improve the stability and efficiency of neural differential equations in stiff biological settings, offering a pathway toward more reliable data-driven modeling of oscillatory biophysical dynamics.

\section{Related Work}
\label{sec:headings}

\subsection{Neural Ordinary Differential Equations and Stiff Dynamics}
Neural ODEs introduced a continuous-depth modeling paradigm where neural networks parameterise the right-hand side of an ODE system. This framework has proven successful in modeling time-series data \citep{NEURIPS2020_4a5876b4}. However, standard Neural ODEs are often trained using higher order solvers (e.g., Runge-Kutta method of order 4 (RK4)) and adjoint methods that assume smooth trajectories and well-conditioned gradients. This becomes problematic in stiff systems, where latent variables evolve on disparate timescales and high Lipschitz constants cause numerical instability \citep{Kim_2021}.

Recent work has begun to address stiffness in learned dynamics via implicit solvers \citep{baker2022proximalimplicitodesolvers}, multi-rate methods \citep{multirate_stiff_ode}, and stiffness-aware training schemes \citep{stiff_aware_ode}. While promising, these studies have largely focused on synthetic benchmarks or mechanical systems with limited application to biophysical dynamics. Moreover, they typically assume dense, noise-free observations; conditions that are rarely satisfied in neuroscience. Our work extends this line by explicitly addressing stiffness in a biophysical benchmark (Hodgkin–Huxley) under noisy observations, where reliability and extrapolation fidelity are critical.

\subsection{Physics-Informed Neural Networks}

Physics-Informed Neural Networks (PINNs) \citep{raissi_pinn} were introduced as a framework to incorporate known physical laws into the learning objective by penalising violations of differential equations. PINNs have been applied to inverse problems \citep{https://doi.org/10.1029/2020WR027642}, partial observations \citep{komatsu2024estimateepidemiologicalparametersgiven}, and multi-physics settings \citep{SUN2024106421}. Despite these advances, PINNs exhibit poor scalability in stiff systems and are known to have poor extrapolation performance outside the training domain \cite{PENWARDEN2023112464}. Residual losses are often dominated by fast variables, while slow variables contribute vanishing gradients, leading to ill-conditioning and slow or unstable convergence \citep{NEURIPS2021_df438e52} \citep{wang2020understandingmitigatinggradientpathologies}.

Hybrid approaches that combine data-driven loss and physics-based regularisation, such as PI-NODEs, attempt to address this by combining data reconstruction and physics residuals \citep{pi-node}. However, these models typically use uniform weighting across residuals, implicitly assuming comparable variable scales. In stiff systems, this assumption fails spectacularly, and optimisation becomes dominated by high-magnitude terms. While these hybrid models improve data efficiency, their sensitivity to variable scaling remains an open challenge in stiff systems. Our work addresses this gap by introducing a scale-aware residual formulation, which normalises residuals across variables, improving conditioning and convergence stability in stiff regimes while enhancing forecasting abilities.

\subsection{Universal Differential Equations}
Universal Differential Equations (UDEs) \citep{rackauckas2021universaldifferentialequationsscientific} extend the idea of hybrid modeling by embedding neural networks or other machine-learned functions into partially known differential equation systems. Unlike standard Neural ODEs, where the entire vector field is parameterised by a black-box model, UDEs preserve known mechanistic components and use data-driven surrogates to fill in uncertain terms. This formulation enables flexible, interpretable, and sample-efficient learning - particularly in scientific and biological systems.

Recent work has demonstrated the efficacy of UDEs in discovering hidden reaction terms in physical systems \citep{gmd-16-6671-2023}, capturing missing physics in climate models \citep{ramadhan2023capturingmissingphysicsclimate}, and learning biological dynamics from limited and noisy measurements \citep{Philipps2024.11.29.626122}.

While conceptually related to PI-NODEs, UDEs differ in that neural components are integrated directly into the symbolic ODE structure, rather than enforced through penalising residuals. This yields interpretable models but introduces challenges in balancing scales across neural and mechanistic terms - a problem also observed in PINNs. Our framework differs by retaining the flexibility of Neural ODEs while applying scale-aware residual constraints to enforce physical consistency. In this way, PI-NODE-SR combines the strengths of UDE-style hybrid modeling with improved training stability for stiff, oscillatory systems. In addition, our scale-aware residual weighting addresses a persistent challenge in both UDEs and PINNs; ensuring balanced learning when system variables evolve across disparate magnitudes and timescales.

\subsection{Machine Learning for Biophysical Modeling}

Neural networks have been increasingly applied to model electro-physiological dynamics, including cardiac and neuronal systems. Several works have attempted to approximate Hodgkin-Huxley-type models with deep learning surrogates and predicting changes in ion channel conductance \cite{ionchangeconductance}. While effective in controlled settings, these approaches typically require long, noise-free recordings, do not address stiffness or extrapolation, and performance degrades outside the training window.

To our knowledge, no prior work has demonstrated stable long-horizon extrapolation of HH dynamics from short, noisy windows using physics-informed methods and explicit lower order solvers. Our work fills this gap by showing that scale-aware residuals enable Physics-Informed NeuralODEs to learn both the frequency and amplitude of oscillations from a single cycle and provide stable extrapolations beyond \SI{100}{\milli\second}.

\section{Methodology}

\subsection{Overview}
We propose a Physics-Informed Neural ODE with Scale-Aware Residuals (PI-NODE-SR) for learning stiff biophysical dynamics from short, noisy time series. Our approach couples a standard data reconstruction loss with a normalised physics residual loss that balances contributions across state variables evolving on disparate timescales. The model receives as input the membrane voltage, gating variables, and time, and outputs temporal derivatives of all states. By enforcing residual consistency in a scale-aware manner, PI-NODE-SR improves training stability, reduces ill-conditioning, and enables long-horizon extrapolation in regimes where conventional Neural ODEs and PI-NODEs fail.

\subsection{The Hodgkin-Huxley Model}

The Hodgkin-Huxley (HH) model describes the nonlinear interaction of the membrane potential of a neuron in response to ionic currents regulated by voltage-dependent gating variables. It is a canonical example of a stiff, nonlinear dynamical system due to the presence of multiple coupled ODEs evolving over disparate timescales.

The membrane potential \( v(t) \) is governed by the following equation:

\begin{equation}
C_m \frac{dv}{dt} = -\bar{g}_{\mathrm{Na}} m^3 h (v - E_{\mathrm{Na}}) - \bar{g}_{\mathrm{K}} n^4 (v - E_{\mathrm{K}}) - \bar{g}_L (v - E_L) + I_{\text{ext}}(t),
\end{equation}

where \( C_m \) is the membrane capacitance, \( \bar{g}_{\mathrm{Na}}, \bar{g}_{\mathrm{K}}, \bar{g}_L \) are the maximal conductance for sodium, potassium, and leak channels respectively, \( E_{\mathrm{Na}}, E_{\mathrm{K}}, E_L \) are the reversal potentials, and \( I_{\text{ext}}(t) \) is the externally applied current. The gating variables \( m(t), h(t), n(t) \in [0, 1] \) represent the fraction of ion channels in the open state.

Each gating variable evolves according to first-order voltage-dependent kinetics:

\begin{equation}
\frac{dx}{dt} = \alpha_x(v)(1 - x) - \beta_x(v)x, \quad \text{for } x \in \{n, m, h\},
\end{equation}

where \( \alpha_x(v) \) and \( \beta_x(v) \) are rate functions that vary nonlinearly with membrane potential. For example, the functions for the potassium activation variable \( n \) are:

\begin{equation}
\begin{aligned}
\alpha_n(v) &= \frac{0.01 (v + 55)}{1 - e^{-0.1 (v + 55)}}, \quad
\beta_n(v) &= 0.125 e^{-0.0125 (v + 65)}.
\end{aligned}
\end{equation}

Similar expressions define \( \alpha_m, \beta_m \) and \( \alpha_h, \beta_h \). Together, these equations form a four-dimensional nonlinear ODE system. 

Strong nonlinearities (\( m^3 h \), \( n^4 \)) and disparate timescales render the HH system stiff, producing oscillatory voltage spikes that are challenging for data-driven models to reproduce. 

In this work, we treat the HH equations as the ground-truth generator of synthetic training data, solved with a stiff solver (Rodas5P) and augmented with noise to mimic experimental conditions. Our goal is to learn the vector field \( f(v, n, m, h, t) \) mapping states and time to the temporal derivatives. The neural model is trained to fit the observed trajectories and minimise physics residuals derived from the aforementioned equations.

\subsection{Neural ODE Architecture}

We model the Hodgkin-Huxley system using a Neural ODE framework. Let \(\mathbf{z}(t) = [v(t), n(t), m(t), h(t)] \in \mathbb{R}^4\) represent the state of the system at time \(t\). We learn a neural network parameterisation of the right-hand side of the ODE:
\begin{equation}
\frac{d\mathbf{z}}{dt} = f_\theta(\mathbf{z}, t)
\end{equation}
where \(f_\theta: \mathbb{R}^5 \rightarrow \mathbb{R}^4\) is a neural network with trainable parameters \(\theta\). Including \(t\) as an explicit input allows the model to learn non-autonomous dynamics and improves extrapolation fidelity.

\medskip

The architecture is a 3-layer MLP with 64 - 128 hidden units per layer, Tanh activations, and layer normalisation for stability. Inputs and outputs are normalised by dataset statistics. Gradients are computed using the QuadratureAdjoint method, which provides accurate sensitivities while remaining compatible with stiff solvers. This enables end-to-end training of the Neural ODE using a composite loss that includes both data and physics-informed terms (detailed in Section~\ref{sec:Loss Function and Physics-Informed Training}).

To simulate trajectories, we solve the learned ODE using a stiff-aware solver. Specifically, we integrate the system using the Rodas5P solver (Rosenbrock-Wanner method), ensuring consistent handling of sharp spikes, over the time interval \([0, T]\) from an initial state \(\mathbf{z}_0\). The predicted trajectory is given by:
\begin{equation}
\hat{\mathbf{z}}(t) = \texttt{ODESolve}(f_\theta, \mathbf{z}_0, [0, T]).
\end{equation}

\subsection{Loss Function and Physics-Informed Training} \label{sec:Loss Function and Physics-Informed Training}

We optimise the Neural ODE by minimising a composite loss function that combines a data reconstruction term with a physics-based residual loss. The total loss is defined as:
\begin{equation}
\mathcal{L}_{\text{total}} = \mathcal{L}_{\text{data}} + \lambda \mathcal{L}_{\text{physics}}
\end{equation}
where \(\lambda > 0\) is a hyperparameter that balances empirical fit and physics consistency.

\subsubsection{Data Loss}
The data loss encourages the model to accurately reproduce the observed membrane voltage and gating dynamics over the training window. Given a set of noisy observations \(\{ \mathbf{z}_i \}_{i=1}^N\), where \(\mathbf{z}_i = [v_i, n_i, m_i, h_i]\), and corresponding neural ODE predictions \(\{ \hat{\mathbf{z}}_i \}_{i=1}^N\), the data loss is defined as:
\begin{equation}
\mathcal{L}_{\text{data}} = \frac{1}{N} \sum_{i=1}^N \| \hat{\mathbf{z}}_i - \mathbf{z}_i \|_2^2
\end{equation}

\subsubsection{Scale-Aware Physics Residual Loss.}

Standard PI-NODEs compute residuals directly from the governing equations, but in stiff systems these terms differ drastically in magnitude; voltage dynamics dominate while gating dynamics vanish. This imbalance yields ill-conditioned gradients. To guide the model toward physically consistent dynamics, we define a residual loss based on the original Hodgkin-Huxley equations. Let \( \mathbf{r}(\mathbf{z}, t) \in \mathbb{R}^4 \) denote the vector of residuals computed by substituting the network output into the governing equations (i.e., the difference between the predicted derivatives and the true Hodgkin-Huxley dynamics).

Given that the components of \(\mathbf{r}(\mathbf{z}, t)\) can vary in scale (e.g. voltage derivatives vs. gating dynamics), we introduce a scale-aware residual formulation.

\begin{equation}
\mathcal{L}_{\text{physics}} = \frac{1}{N} \sum_{i=1}^N \sum_{j=1}^4 \left( \frac{r_{ij}}{s_j} \right)^2,
\end{equation}

where \( r_{ij} \) is the \( j \)-th component of the residual in sample \( i \), and \( s_j \) is a scaling factor associated with each variable. In our experiments, we set \( s_j \) as the empirical standard deviation of the derivative of the target variable in the training window.

This normalisation ensures that each variable contributes proportionally to the gradient, preventing the small-magnitude terms of the gating dynamics from being overwhelmed by larger component of the voltage dynamics. The result is improved convergence, more balanced learning, and enhanced extrapolation stability in stiff dynamical regimes. This formulation is the core novelty of PI-NODE-SR; it stabilises training under stiff dynamics and enables convergence within practical iteration budgets.

\subsection{Training and Data Generation}
HH trajectories are simulated with Rodas5P ~\cite{rackauckas2017differentialequations} over \SI{100}{\milli\second}, with external current pulses \( I_{\text{ext}}(t) \) to elicit spiking. A \SI{14.68}{\milli\second} segment (\(t \in [0,14.68]\)) is used for training, with full-state observations \( [v(t), n(t), m(t), h(t)] \) sampled at \SI{0.01}{\milli\second} resolution (1468 data points).

To mimic experimental conditions, we augment each channel with a deterministic Ornstein–Uhlenbeck process:

\begin{equation}
\tilde{z}(t) = z(t) + \xi(t), \quad \text{where } \xi(t + \Delta t) = \xi(t) - \theta \xi(t) \Delta t + \sigma \sqrt{\Delta t} \, \mathcal{N}(0, 1).
\end{equation}

We set \(\theta = 5.0\) and \(\sigma = 0.5\) to produce smooth, bounded perturbations that remain consistent across training runs. This preserves the deterministic structure of the training ODE while modeling biologically plausible signal variability. Models are trained using the Adam Optimiser with a learning rate of \(\eta = 10^{-3}\) for up to 5000 epochs.

\subsection{Evaluation} \label{subsec:Evaluation}
To assess short-term fidelity and long-term extrapolation, we evaluate predictions on (\(t \in [14.68, 100]\)) using both trajectory and spike-based metrics:

\begin{itemize}
  \item \textbf{Root Mean Squared Error (RMSE)} between predicted and ground-truth trajectories for membrane potential and gating variables.
  
  \item \textbf{Amplitude Preservation Error (\%)} computed as the relative deviation in the peak and trough values of voltage and gating variables over the extrapolation window.
  
  \item \textbf{Phase Error (ms)} defined as the average absolute deviation in spike timing (peak membrane potential events) between prediction and ground truth.
  
  \item \textbf{Inter-Spike Interval (ISI) Error (ms)} quantifying rhythm preservation via differences in the timing between successive action potentials.

  \item \textbf{Spike Detection F1 Score} representing the harmonic mean of spike precision and recall.
\end{itemize}

These metrics complement trajectory error by quantifying biologically critical properties; spike timing, rhythm and oscillation shape. Additionally, we visualise extrapolated traces, gating dynamics, and phase portraits to assess qualitative stability. We compare our method against two widely used learning-based baselines:
\begin{itemize}
  \item \textbf{Vanilla Neural ODE:} A non-physics-informed neural ODE trained solely on data loss \(\mathcal{L}_{\text{data}}\).
  \item \textbf{Vanilla PINN:} A physics-informed model trained using the HH equations, without scale-aware normalisation.
\end{itemize}

All models are trained on the same noisy input and use the same network architecture and solver to ensure fair comparison.

\section{Results and Analysis}

We evaluate PI-NODE-SR on the HH system. Models are trained on a single oscillation (\SI{14.68}{\milli\second}) and tested on a \SI{100}{\milli\second} extrapolation horizon, corresponding to $\approx$7 spike cycles. We compare against two baselines: (i) a vanilla PINN trained solely via residual minimisation, and (ii) a vanilla Neural ODE trained only on supervised data. Performance is assessed in terms of short-term reconstruction, long-term stability, spike timing, phase portraits, and gating variable fidelity.

\subsection{Failure Modes of Vanilla NeuralODEs}
\label{sec:vanilla-neuralode}

A vanilla Neural ODE trained on short-window supervised data fails to generalise beyond the training domain. Predictions are integrated with \texttt{Rodas5P} (stiff solver), \texttt{Tsit5} (adaptive Runge–Kutta), and \texttt{Heun} (explicit 2nd-order), with results displayed in Figure~\ref{fig:vanilla-neuralode-results}. Across solvers, we observe the following:

\begin{figure}
    \centering    \includegraphics[width=0.95\textwidth]{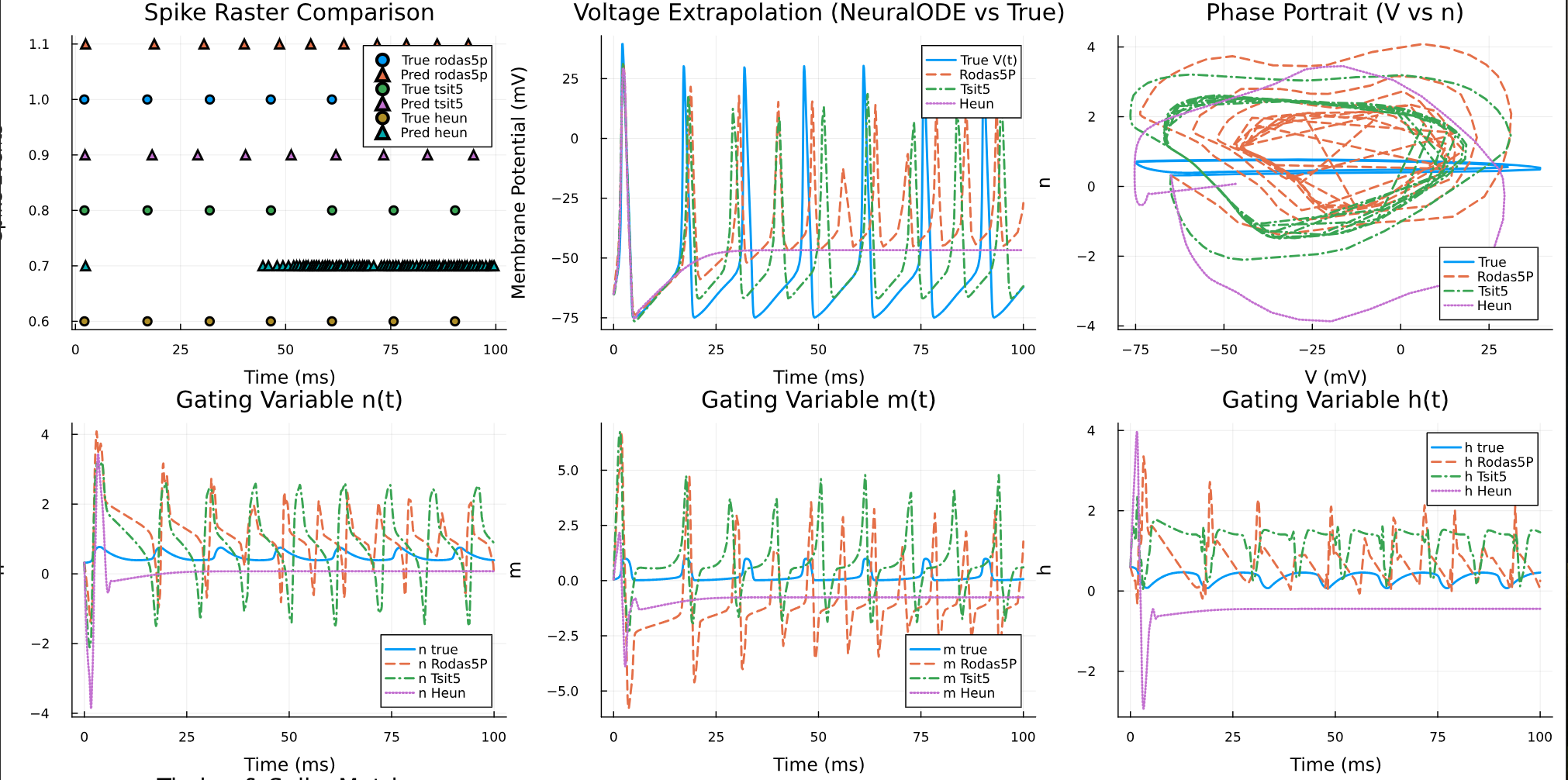}
    \caption{\textbf{Extrapolation behaviour of vanilla NeuralODE models trained on Hodgkin–Huxley dynamics using different solvers}. Top: spike raster, voltage $V(t)$, and phase portrait. Bottom: gating variable predictions ($n$, $m$, $h$).}
    \label{fig:vanilla-neuralode-results}
\end{figure}

\begin{itemize}
    \item \textbf{Temporal drift}: Spike timings initially align, but phase errors accumulate rapidly, leading to millisecond-scale misalignments. No generalisation beyond the training data is observed when using \texttt{Heun's method}, showing the complete failure of explicit, lower order solvers.
    \item \textbf{Voltage degradation}: Spikes attenuate over time. With \texttt{Heun}, the model collapses to a low-amplitude regime; with \texttt{Tsit5}, oscillations fade; \texttt{Rodas5P} maintains spiking longer but shows jitter and decay.
    \item \textbf{Attractor instability}: The phase portrait ($V$–$n$) reveals distorted, divergent trajectories, proving that learned dynamics do not converge to the HH system's limit cycle across all 3 solvers.
    \item \textbf{Gating variable explosion}: All variants eventually violate the biophysical constraints of the gating variables $[0,1]$. The predicted values for $m$ and $h$ reach magnitudes above $5$, especially with \texttt{Heun}, indicating unbounded growth and numerical and representational instability.
\end{itemize}

\vspace{2.5em}
These results highlight the sensitivity of NeuralODEs to solver choice and the inability of data-only training to sustain stiff oscillations. The choice of numerical solver significantly affects the stability and long-term accuracy of the model, with explicit solvers (e.g., \texttt{Heun}) performing notably worse than stiff solvers like \texttt{Rodas5P}. Without inductive bias, normalisation, structural constraints, or physics-based regularisation, the models overfit short-term behavior and exhibit poor generalisation.

\subsection{Failure Modes of Vanilla PINNs}
\label{sec:vanilla-pinn}

We test a residual-based PINN trained on the same short window. Unlike Neural ODEs, the PINN minimises ODE residuals without scale-aware weighting. This setup constitutes a baseline test of the model's ability to infer sustained dynamics solely from minimising short-term physics error.

\begin{figure}[H]
  \centering
  \includegraphics[width=\textwidth]{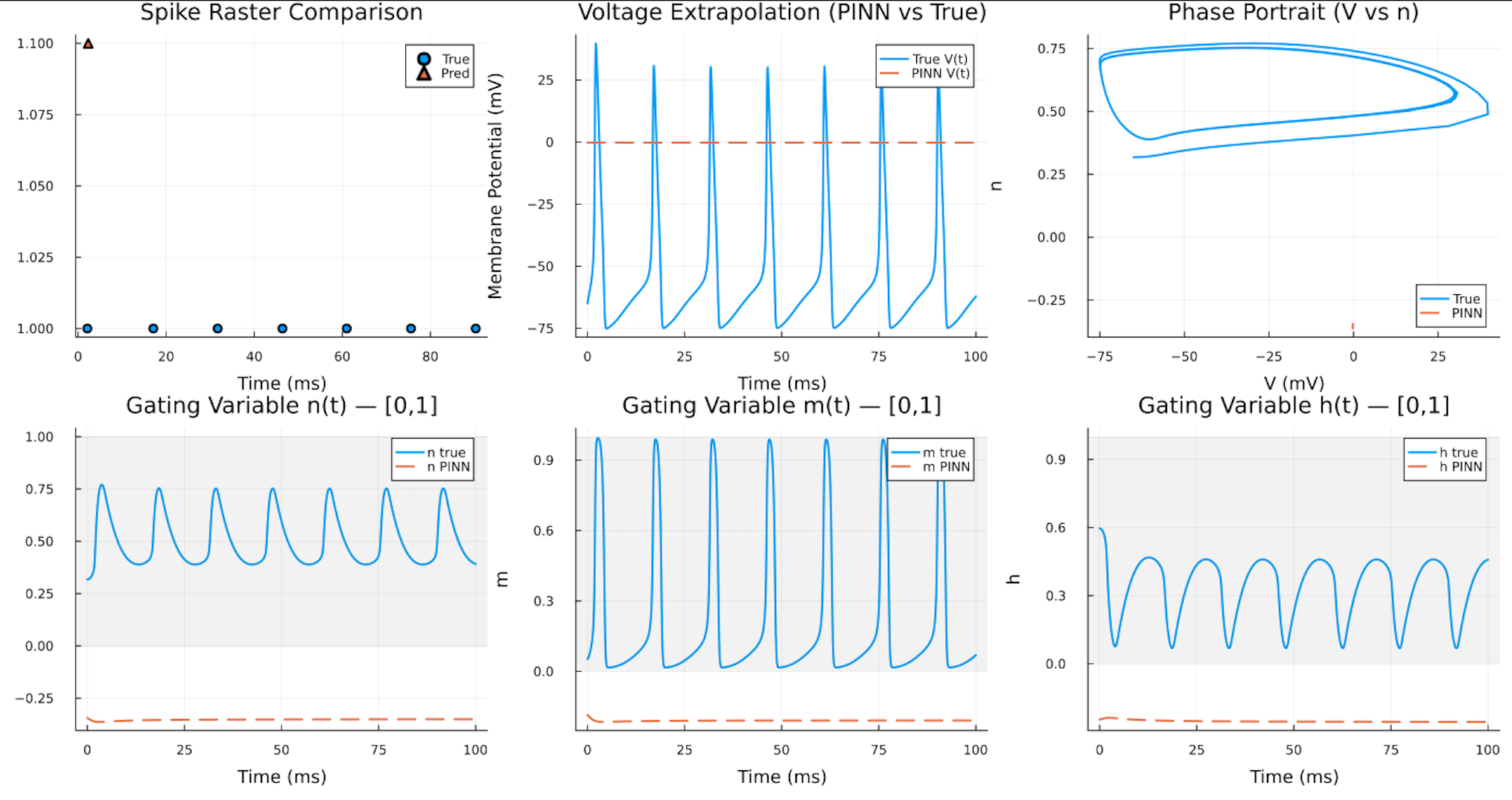}
  \caption{
    \textbf{Vanilla PINN performance on Hodgkin–Huxley dynamics.}
    \emph{Top Left:} Spike raster shows zero predicted spikes beyond training.
    \emph{Top Middle:} Voltage traces collapse to a flat resting potential.
    \emph{Top Right:} Phase portrait shows no learned oscillatory structure.
    \emph{Bottom Row:} Gating variables remain flat and outside biological bounds.
  }
  \label{fig:vanilla_PINN_results}
\end{figure}

As shown in Figure~\ref{fig:vanilla_PINN_results}, the PINN fails to reproduce the long-term spiking behavior of the HH system. While it may begin to locally minimise the ODE residual on the short training window, the learned solution exhibits the following:

\begin{itemize}[itemsep=0.5pt, topsep=0pt]
    \item \textbf{Collapsed dynamics:} The predicted voltage quickly flattens to a subthreshold resting state with no further spiking, failing to initiate any further action potentials.
    \item \textbf{No limit-cycle formation:} The phase portrait ($V$–$n$) confirms that the model converges to a near-fixed point rather than reproducing the characteristic limit cycle of the system.
    \item \textbf{Static gating variables:} Predicted values of $n(t)$, $m(t)$, and $h(t)$ stay close to their initialisation and do not evolve in time. They remain outside the biophysical $[0, 1]$ interval and show no oscillatory structure.
\end{itemize}

This behavior is consistent with known limitations of vanilla PINNs \cite{WANG2022110768} in modeling nonlinear oscillatory systems. Even when residual loss is effectively minimised in a narrow temporal region, PINNs lack the inductive bias to generate sustained rhythms or attractor dynamics. The mismatch between local residual minimisation and global trajectory structure is especially severe in stiff systems like Hodgkin–Huxley, where long-range dependencies dominate. Thus, we see that the PINN does not reproduce the oscillatory limit cycle and instead predicts convergence to a fixed point.

In general, these results highlight a key insight: \textbf{residual minimisation alone is insufficient in  stiff oscillatory systems to guarantee long-term predictive stability}; the PINN reduces local residuals but fails to recover global limit-cycle behaviour.

\subsection{Summary of Baseline Limitations}

Both baselines expose complementary weaknesses. Vanilla Neural ODEs overfit short-term data and degrade under extrapolation, while vanilla PINNs collapse despite residual minimisation. Neither approach balances variable scales, leading to unstable training and poor long-horizon dynamics. These failures motivate PI-NODE-SR, which combines physics residuals with scale normalisation to better stabilise learning.

\subsection{PI-NODE-SR: Stable Long-Horizon Dynamics}

We train PI-NODE-SR on a single \SI{14.68}{\milli\second} oscillation using Heun’s method as the explicit integration scheme. Despite Heun’s unsuitability for stiff systems in isolation, its combination with scale-aware residuals yields more stable training and accurate long-term forecasts, with qualitatives behaviours summarised in Table~\ref{tab:model-comparison}.

\begin{table}[H]
\centering
\caption{Qualitative comparison of model behaviors over long-term Hodgkin–Huxley extrapolation.}
\label{tab:model-comparison}
\begin{tabular}{lcccc}
\toprule
\textbf{Model} & \textbf{Spike F1} & \textbf{Phase Portrait} & \textbf{Gating Variables} & \textbf{Stability} \\
\midrule
Vanilla PINN & 0.00 & No (collapsed) & No (flat / invalid) & Fails early \\
Vanilla NeuralODE & 0.40–0.44 & Poor (divergent) & Partially (unstable) & Degrades \\
\textbf{PI-NODE-SR (Ours)} & \textbf{0.98} & Yes (limit-cycle) & Yes (biophysical) & Stable \\
\bottomrule
\end{tabular}
\end{table}

\subsubsection{Voltage Trajectory and Spike Timing}

\begin{figure}[H]
    \centering   \includegraphics[width=0.95\textwidth]{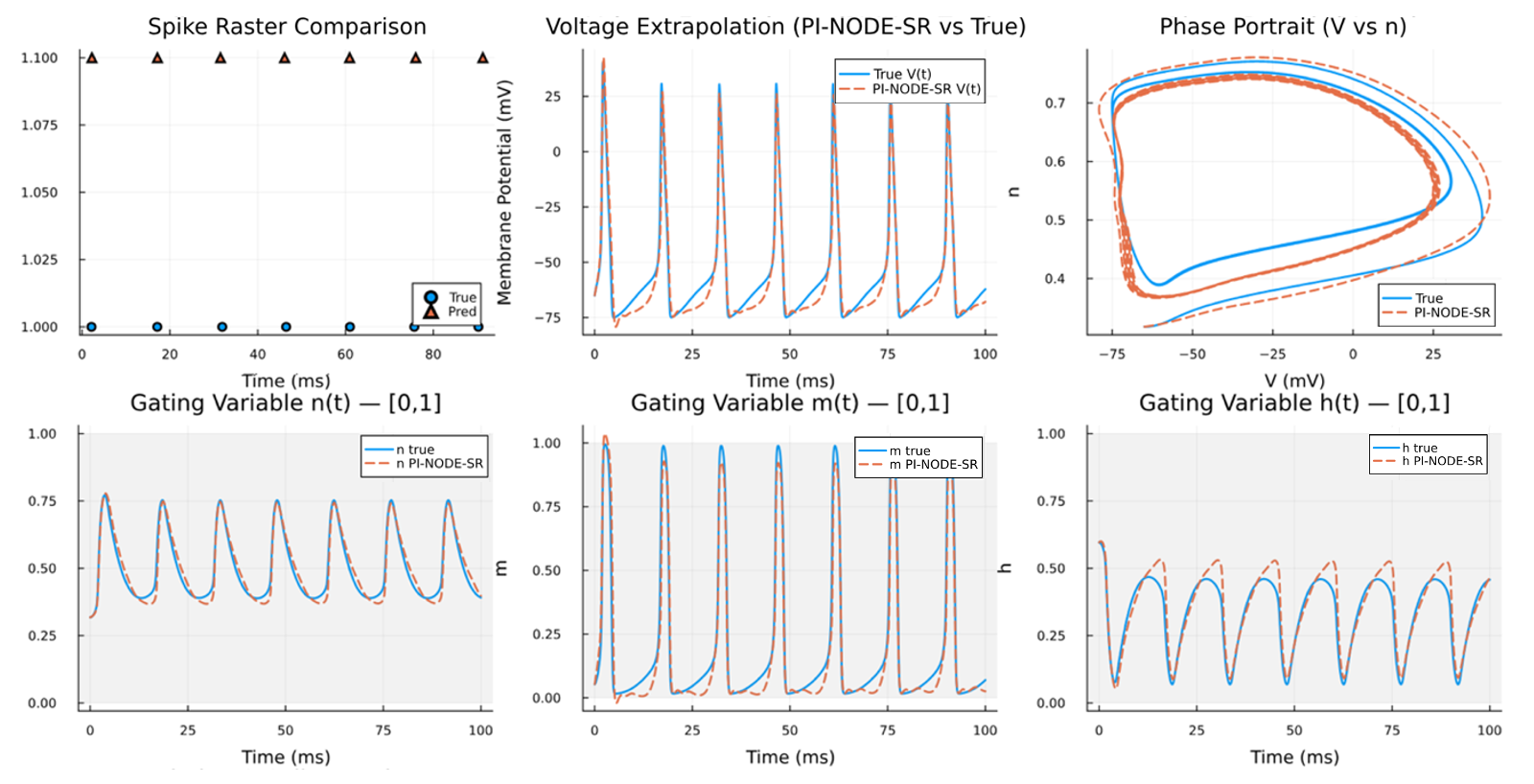}
    \caption{\textbf{PI-NODE-SR extrapolation results on Hodgkin-Huxley dynamics}. Top left: spike raster comparison; Top center: predicted vs. ground truth membrane potential $V(t)$; Top right: phase portrait ($V$–$n$); Bottom: gating variable trajectories ($n$, $m$, $h$).}
    \label{fig:pinode-sr-results}
\end{figure}

Figure~\ref{fig:pinode-sr-results} shows that PI-NODE-SR closely reproduces HH voltage dynamics over \SI{100}{\milli\second}. Action potentials match in both amplitude and timing, with:

\begin{itemize}
    \item \textbf{Phase shift:} $\Delta\phi = 0.27$ ms
    \item \textbf{Frequency error:} 0.02 ms
    \item \textbf{Spike F1-score:} 0.98
\end{itemize}

\vspace{-1.5em}
This demonstrates that the model captures oscillatory frequency and preserves spike structure over long horizons with high fidelity.

\subsubsection{Gating Variable Dynamics}

The predictions for $n(t)$ and $m(t)$ overlap the ground truth, while $h(t)$ is slightly overestimated in amplitude; the correct oscillation frequency is maintained throughout the extrapolation. All gating variables remain within the biophysical range, showing that PI-NODE-SR respects constraints absent in baselines while learning both the fast and slow time-scale components of the HH system.

\subsubsection{Phase Portrait and System Dynamics}

The phase portrait ($V$–$n$) confirms that PI-NODE-SR reconstructs a functionally accurate version of the HH systems limit cycle. The learned attractor is slightly smoothed, with minor deviations in orbit width and curvature, but qualitatively accurate, preserving the oscillatory geometry absent in baseline models.

\subsubsection{Robustness Across Runs}

Across 10 random initialisations, PI-NODE-SR consistently outperformed baselines. Convergence occurred between 3,000 and 5,000 iterations, with variance in early training stability but consistent long-term error reduction. Although initialisation sensitivity remains, scale-aware residuals seem to prevent collapse and yield more stable extrapolations.

\subsubsection{Key Findings}

\begin{itemize}
    \item Purely data-driven Neural ODEs drift and degrade; PINNs collapse.
    \item Scale-aware residuals restore balance across variables and aid towards more stable training.
    \item PI-NODE-SR achieves accurate voltage, gating, and attractor dynamics over long horizons, despite using a low-order explicit solver.
    \item Solver-loss synergy (Heun + scale-aware residuals) is critical: Heun alone fails, but in combination enables more robust learning.
\end{itemize}

\subsection{Model Comparison Summary}
We benchmark three families of approaches - vanilla NeuralODEs, vanilla PINNs, and our proposed PI-NODE-SR - on the Hodgkin–Huxley (HH) system. Across all metrics, PI-NODE-SR achieves the only stable long-term extrapolation, maintaining consistent magnitudes and oscillation frequencies. Vanilla NeuralODEs capture the first oscillation but collapse into flat or divergent dynamics depending on solver choice and normalisation, with RMSE(V) exceeding \SI{297.98}{\milli\volt} in extreme cases. In contrast, PI-NODE-SR maintains spike alignment within \SI{0.27}{\milli\second} phase drift, preserves amplitudes within \SI{0.62}{\percent} and achieves an F1-score of 0.98, reflecting near-perfect recovery of oscillatory structure as shown in Table~\ref{tab:voltage-metrics}. Gating variable RMSEs remain below 0.1 across \SI{100}{\milli\second} extrapolation horizons as shown in Table~\ref{tab:gating-rmse}. These results establish that scale-aware residuals, time-aware dynamics, and physics-informed regularisation are all required to move beyond short-term fitting toward biologically plausible long-range prediction.

\vspace{-1.2em}
\section{Ablation Study}
To understand the contribution of each design choice in our PI-NODE-SR, we conduct an ablation study systematically by removing or replacing individual components of the framework and compare against baseline models with normalisation. Tables~\ref{tab:voltage-metrics} and ~\ref{tab:gating-rmse} summarises the performance of each variant using metrics defined in Section~\ref{subsec:Evaluation}. All models are trained on identical simulation data (one oscillation of \SI{14.68}{\milli\second}) and evaluated on long-term extrapolation (\SI{14.68}{\milli\second}-\SI{100}{\milli\second}). This allows us to assess whether normalisation alone is sufficient to recover long-term fidelity in existing baseline models or whether deeper architectural changes are required.

\vspace{-0.6em}
\begin{table}[H]
\centering
\caption{Spike-related voltage metrics computed on extrapolated membrane voltage. Lower is better ($\downarrow$), except for F1 Score where higher is better ($\uparrow$). A dash (-) indicates failure cases where no meaningful metric could be computed due to divergence or complete absence of valid spikes}
\label{tab:voltage-metrics}
\resizebox{\textwidth}{!}{%
\begin{tabular}{lcccccc}
\toprule
\textbf{Model Variant} 
& \textbf{RMSE (V)} $\downarrow$
& \textbf{Phase Drift (ms)} $\downarrow$
& \textbf{ISI Error (ms)} $\downarrow$
& \textbf{Amp. Deviation (mV)} $\downarrow$
& \textbf{Amp. Pres. Error (\%)} $\downarrow$
& \textbf{Spike F1 Score} $\uparrow$ \\
\midrule
Full PI-NODE-SR (ours)              & \textbf{3.12} & \textbf{0.27} & \textbf{0.02} & \textbf{0.05} & \textbf{0.62\%} & \textbf{0.98} \\
\midrule
\emph{-- No physics loss}        & 22.40 & - & - & - & - & 0.00 \\
\emph{-- Backsolve adjoint}      & 23.01 & - & - & - & - & 0.00 \\
\emph{-- No noise}               & 11.31 & 0.37 & 0.13 & 1.12 & 2.57\% & 0.96 \\
\emph{-- $\lambda = 0.5$}        & 15.51 & 0.80 & 0.35 & 10.27 & 22.59\% & 0.91 \\
\emph{-- $\lambda = 2.5$}        & 12.88 & 0.31 & 0.09 & 0.96 & 2.8\% & 0.96 \\
\midrule
\multicolumn{7}{l}{\textit{Vanilla NeuralODE}} \\
\emph{-- Rodas5p with normalisation}     & 18.44 & 0.71 & 1.13 & 22.21 & 43.40\% & 0.44 \\
\emph{-- Rodas5p without normalisation}  & 42.20 & - & - & - & - & 0.00 \\
\emph{-- Tsit5 with normalisation}       & 16.72 & 4.18 & 0.44 & 16.78 & 23\% & 0.40 \\
\emph{-- Tsit5 without normalisation}    & 297.98 & - & - & - & - & 0.00 \\
\emph{-- Heun with normalisation}        &        & 34.71 & 2.34 & 2.39 & - & 0.44 \\
\emph{-- Heun without normalisation}     & 38.55 & - & - & - & - & 0.00 \\
\midrule
\multicolumn{7}{l}{\textit{Vanilla PINN}} \\
\emph{-- with normalisation}             & 69.62 & - & - & - & 74.96\% & 0.00 \\
\emph{-- without normalisation}          & 61.08 & - & - & - & 99.47\% & 0.00 \\
\bottomrule
\end{tabular}%
}
\end{table}

\begin{table}[H]
\centering
\caption{RMSE of gating variables (\textit{n}, \textit{m}, \textit{h}) on long-range predictions from 100 ms to 1000 ms. Lower is better (↓). RMSE for membrane voltage is reported in Table~\ref{tab:voltage-metrics}.}
\label{tab:gating-rmse}
\begin{tabular}{lccc}
\toprule
\textbf{Model Variant} & \textbf{RMSE (n)} ↓ & \textbf{RMSE (m)} ↓ & \textbf{RMSE (h)} ↓ \\
\midrule
Full PI-NODE-SR (ours)              & \textbf{0.02} & \textbf{0.02} & 0.09 \\
\midrule
\emph{-- No physics loss}        & 0.14 & 0.28 & 0.15 \\
\emph{-- Backsolve adjoint}      & 0.15 & 0.26 & 0.17 \\
\emph{-- No noise}               & 0.03 & 0.16 & \textbf{0.04} \\
\emph{-- $\lambda = 0.5$}        & 0.06 & 0.18 & 0.07 \\
\emph{-- $\lambda = 2.5$}          & 0.03 & 0.16 & \textbf{0.04} \\
\midrule
Vanilla NeuralODE \\
\emph{-- Rodas5p with normalisation}     & 0.42 & 12.39 & 2.42 \\
\emph{-- Rodas5p without normalisation}  & 1.06 & 0.62 & 0.49 \\
\emph{-- Tsit5 with normalisation}     & 0.81 & 0.50 & 0.27 \\
\emph{-- Tsit5 without normalisation}  & 0.89 & 2.54 & 15.78 \\
\emph{-- Heun with normalisation}     & 0.91 & 0.29 & 0.32 \\
\emph{-- Heun without normalisation}  & 0.63 & 0.90 & 0.31 \\
\midrule
Vanilla PINN \\
\emph{-- with normalisation}     & 0.29 & 0.43 & 0.16 \\
\emph{-- without normalisation}  & 0.72 & 0.31 & 0.18 \\
\bottomrule
\end{tabular}
\end{table}

\subsection{Effect of Physics Loss Weight \texorpdfstring{$\lambda$}{lambda}}

The physics loss weight $\lambda$ governs the balance between data supervision and Hodgkin–Huxley residual enforcement.

\begin{figure}[h]
    \centering
    \subfloat[$\lambda = 0.0$ (No physics loss)\label{fig:lambda0}]{
        \includegraphics[width=0.31\textwidth]{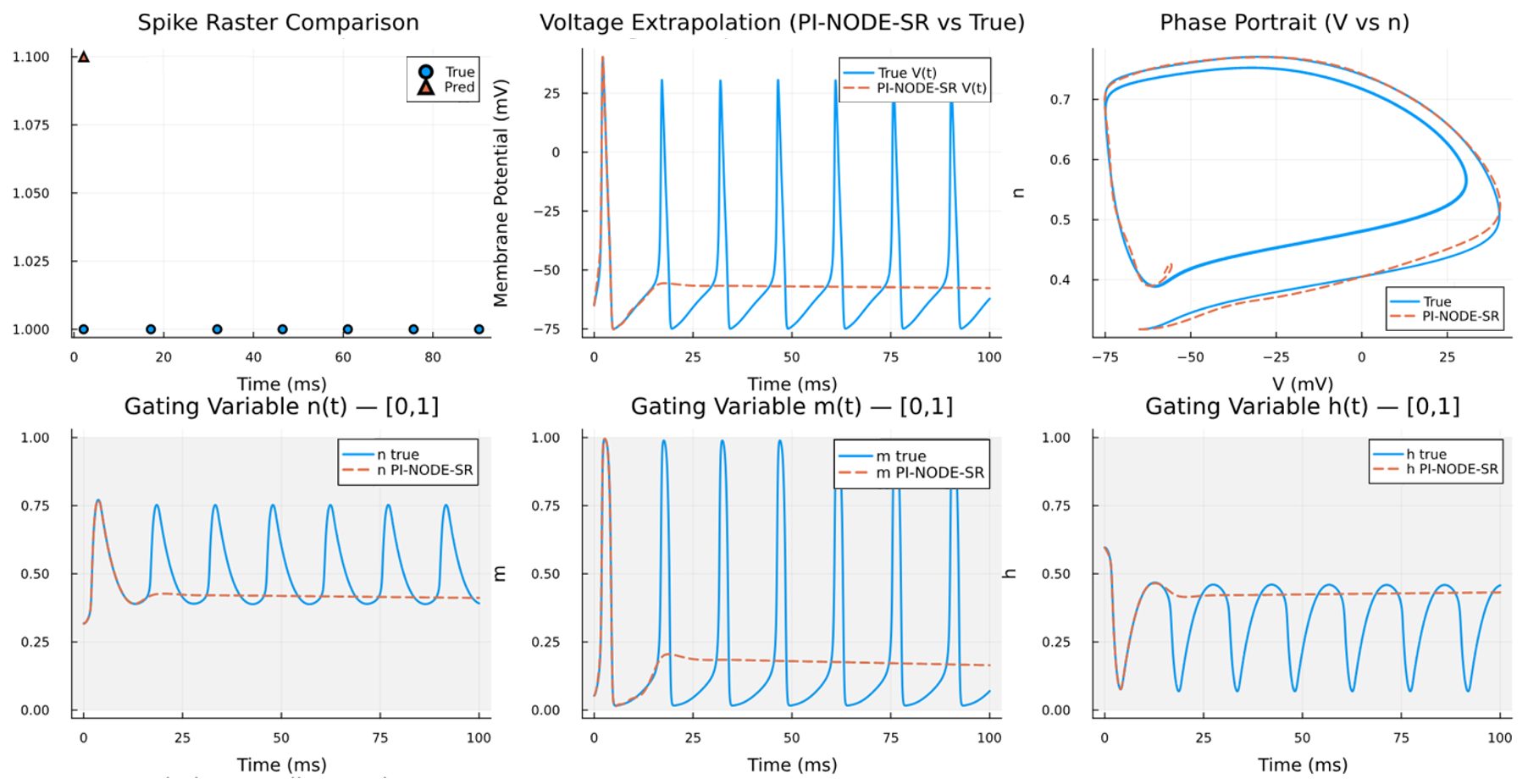}
    }
    \hfill
    \subfloat[$\lambda = 0.5$ (Lower lambda scaling)\label{fig:lambda05}]{
        \includegraphics[width=0.31\textwidth]{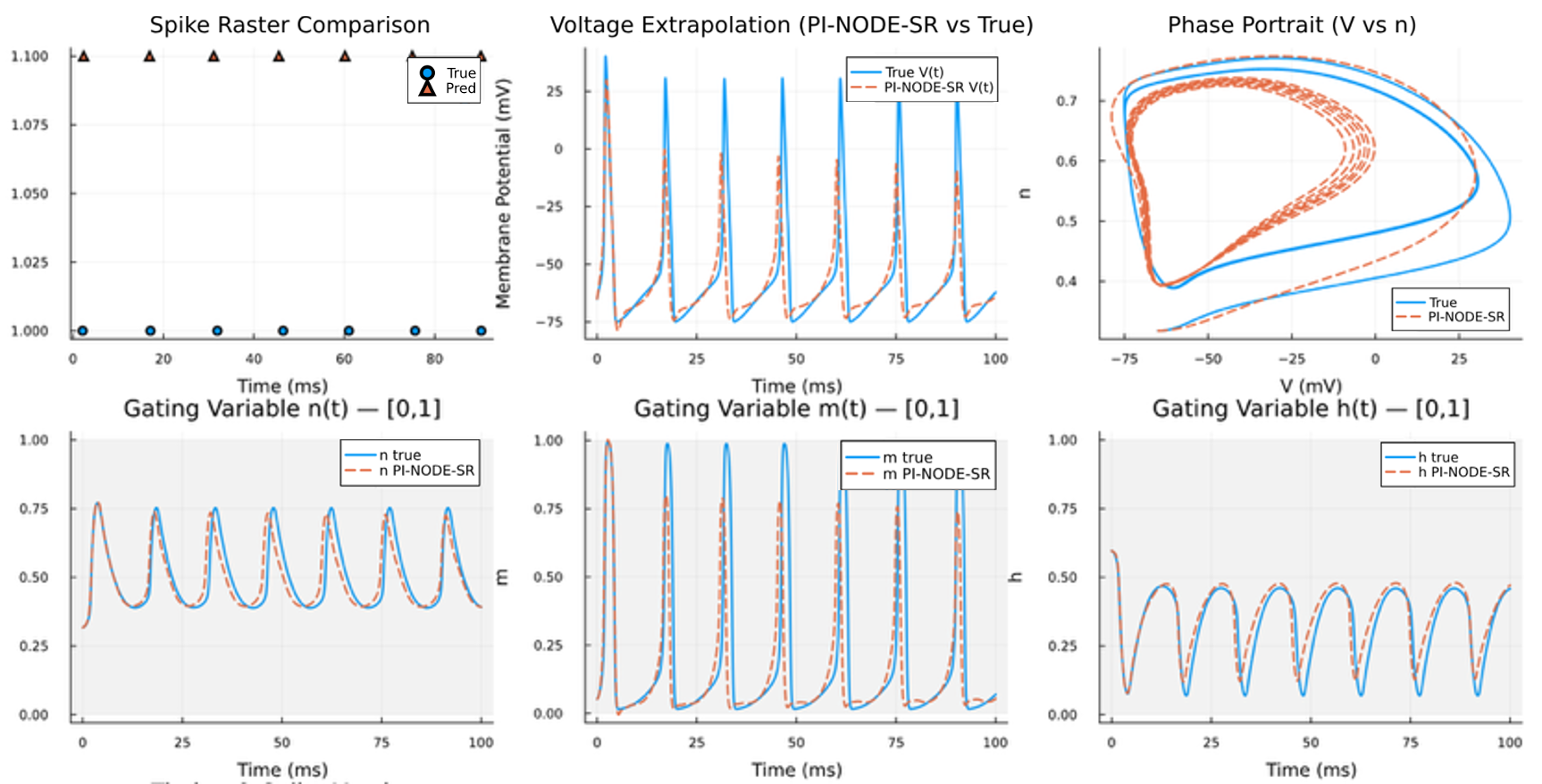}
    }
    \hfill
    \subfloat[$\lambda = 2.5$ (Higher lambda scaling)\label{fig:lambda25}]{
        \includegraphics[width=0.31\textwidth]{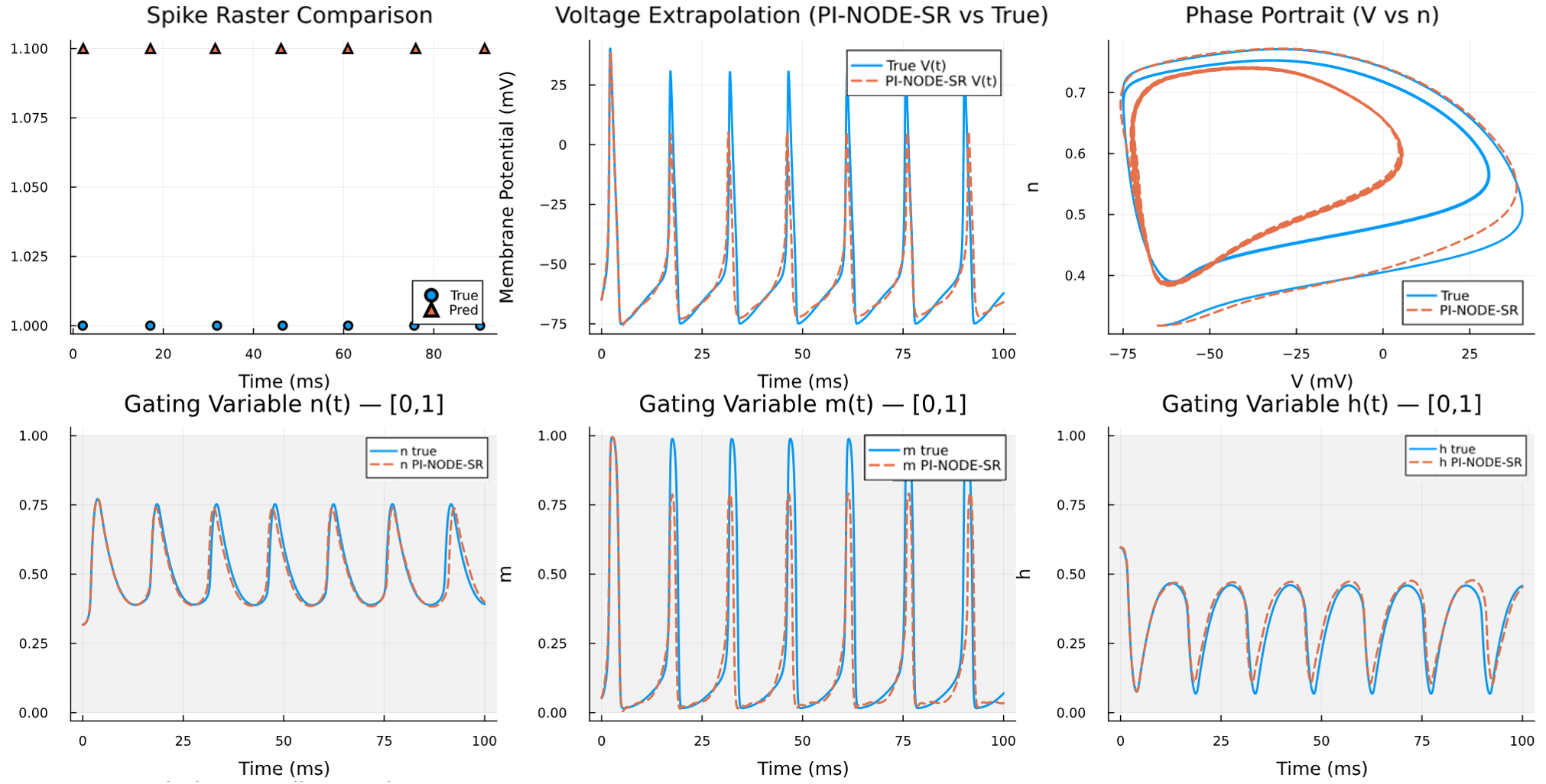}
    }
    \caption{
    \textbf{Impact of physics loss coefficient $\lambda$ on long-term extrapolation}. Increasing $\lambda$ improves structural fidelity, but overly high values smooth out action potentials and slightly degrade timing/precision. $\lambda=1.5$ achieves the best trade-off.
    }
    \vspace{1mm}
    \label{fig:lambda-ablation}
\end{figure}

\begin{itemize}[itemsep=0.5\baselineskip, topsep=0pt, parsep=0pt, partopsep=0pt]
    \item \textbf{$\lambda$ = 0.0} (no physics loss): The model degenerates into a purely data-driven NeuralODE, overfitting the training window. Spiking activity collapses entirely (F1 = 0.00), gating RMSEs exceed 0.14, spiking up to 0.28 for the m-gate, and the voltage trace flattens to a subthreshold state (Figure~\ref{fig:lambda0} ).
    \item \textbf{$\lambda$ = 0.5} (weak regularisation): Oscillations emerge but drift accumulates, with a 0.80 ms phase shift and 22.6\% amplitude error. Spike F1 falls to 0.91 (Figure~\ref{fig:lambda05}), indicating that physics loss is insufficiently strong to stabilise long-range dynamics.
    \item \textbf{$\lambda$ = 2.5}: (strong regularisation): Structural fidelity improves; gating trajectories match ground truth and spikes persist. However, over-smoothing dampens voltage amplitudes  and slightly reduces precision (F1 = 0.96, Figure~\ref{fig:lambda25}).
    \item Best trade-off: Moderate $\lambda$ ($\approx 1.5$) yields sharp spikes, stable gating dynamics, and minimal drift.
\end{itemize}

\vspace{1.5em}
Our finidings shows that the physics loss is indispensable to our proposed model. Moderate values of $\lambda$ can introduce helpful inductive bias without over-smoothing; too little yields collapse and too much suppresses sharp transients.

\subsection{Adjoint method}

\begin{figure}
    \centering   \includegraphics[width=0.95\textwidth]{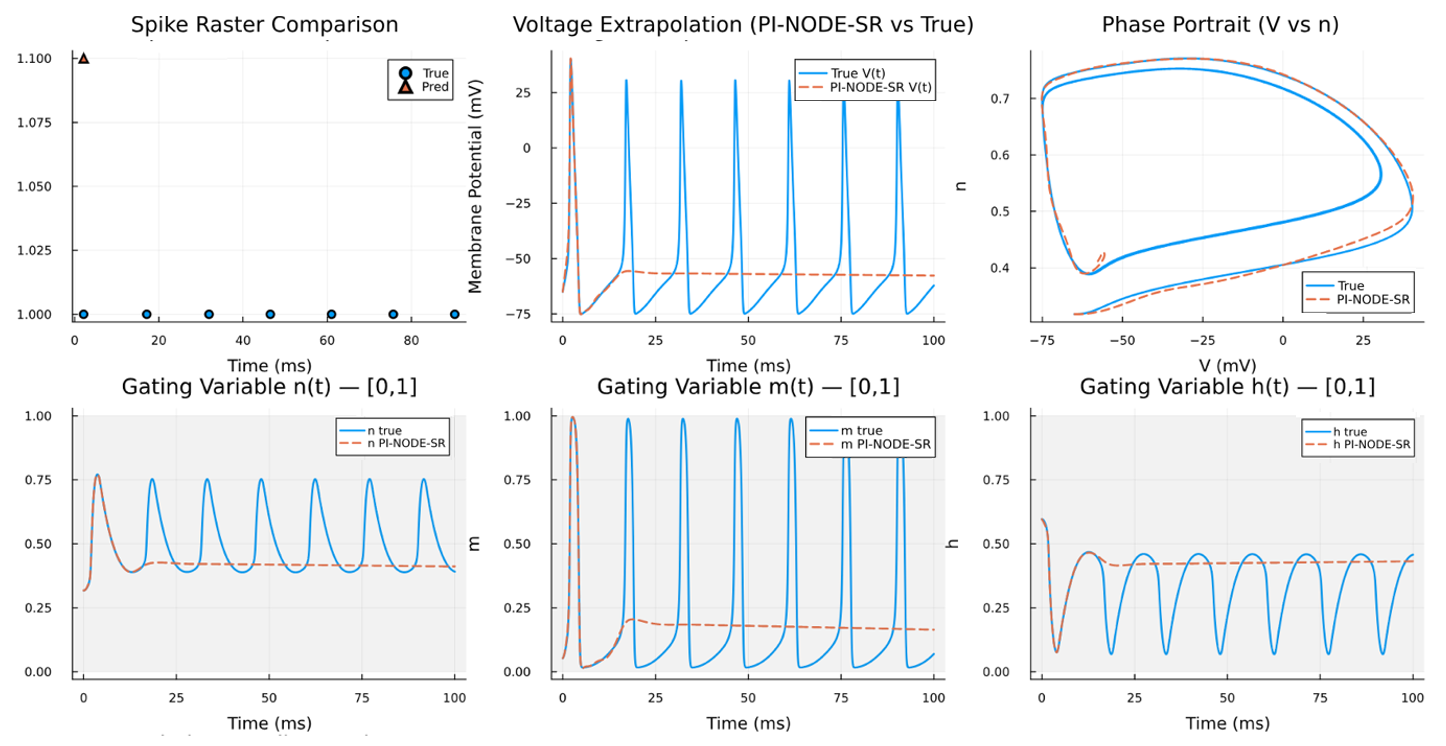}
    \caption{\textbf{PI-NODE-SR with backsolve adjoint extrapolation results on Hodgkin-Huxley dynamics.} Top left: spike raster comparison; Top center: predicted vs. ground truth membrane potential $V(t)$; Top right: phase portrait ($V$–$n$); Bottom: gating variable trajectories ($n$, $m$, $h$).}
    \label{fig:pinode-backsolve-results}
\end{figure}

Switching from the interpolating adjoint to the backsolve adjoint destabilises training. Despite identical architecture and solver settings, the backsolve variant fails to extrapolate beyond the training window, collapsing to flat voltage traces, as shown in Figure~\ref{fig:pinode-backsolve-results} . Quantitatively, RMSE(V) rises to \SI{23.01}{\milli\volt}, gating RMSEs severely degrade in gating variable accuracy (RMSE(n, m, h) = \num{0.15}, \num{0.26}, \num{0.17}), and F1-score falls to 0.00 as shown in Tables~\ref{tab:voltage-metrics} and ~\ref{tab:gating-rmse}. Valid action potentials were not generated after the initial cycle, and both phase portrait and spike raster plots confirm the absence of temporal or structural fidelity. This fragility arises because backsolve adjoints need to integrate gradients backward through stiff spike dynamics, where rapid transients induce exploding/vanishing sensitivities. Interpolating adjoints, by contrast, provide a smoother spline-based trajectory for backpropagation, enabling stable supervision through sharp spikes.
Thus, \textbf{gradient path stability is crucial in stiff systems}; interpolating adjoints are substantially more robust than backsolve adjoints; the latter may compromise expressive capacity, even when architecture and loss remain unchanged.

\subsection{Role of Biophysical Noise}

We test PI-NODE-SR with and without deterministic Ornstein–Uhlenbeck-like noise injected during training.

With noise, the spikes align within \SI{0.27}{\milli\second} drift, amplitudes are better preserved (0.62\% error), and phase portraits overlap true orbits. Gating variables $n$ and $m$ are reconstructed with RMSE < \num{0.02}, maintaining stable periodicity and correct phase relationships over extended extrapolation.

\begin{figure}
    \centering   \includegraphics[width=0.95\textwidth]{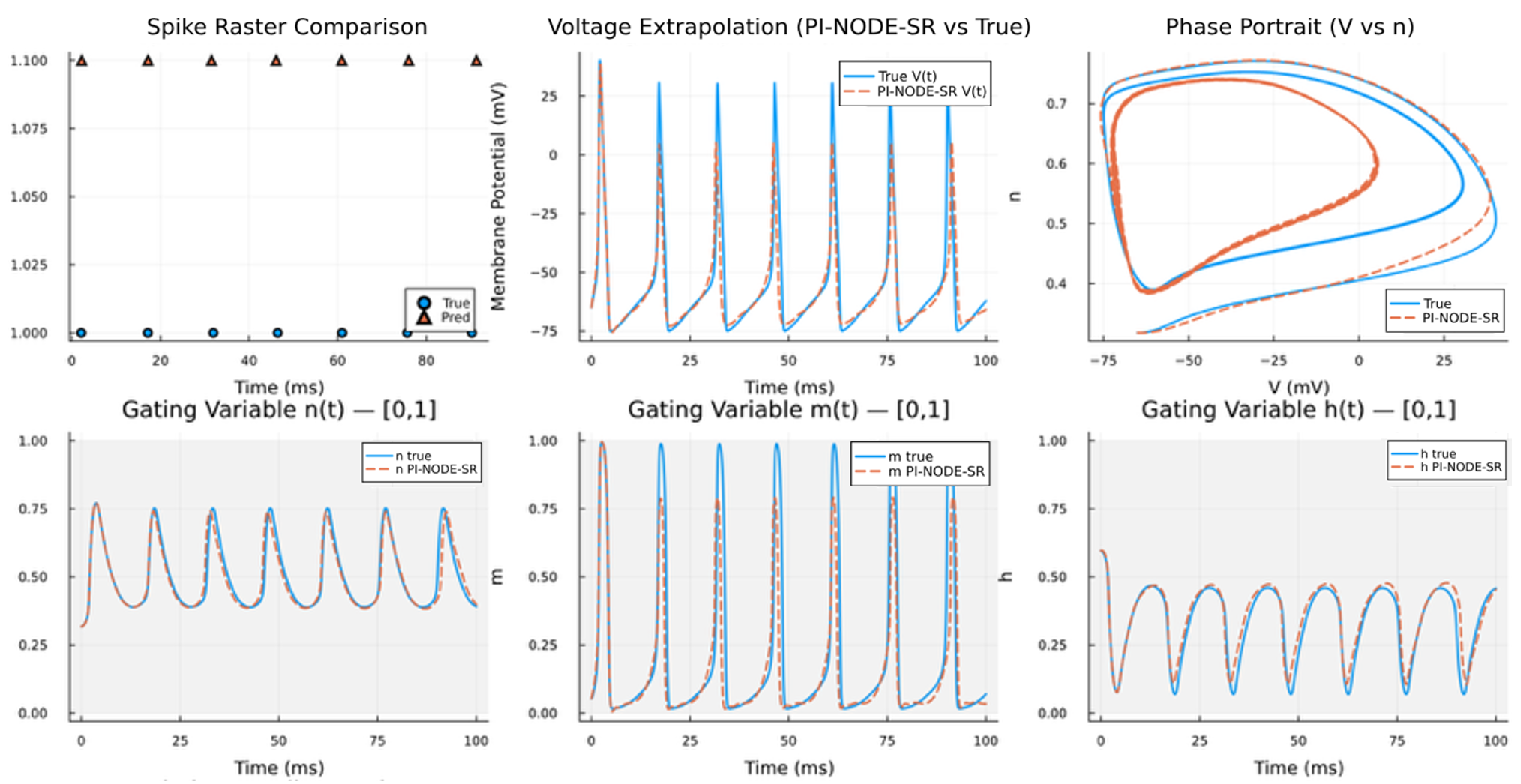}
    \caption{\textbf{PI-NODE-SR without noise on Hodgkin-Huxley dynamics}. Top left: spike raster comparison; Top center: predicted vs. ground truth membrane potential $V(t)$; Top right: phase portrait ($V$–$n$); Bottom: gating variable trajectories ($n$, $m$, $h$).}
    \label{fig:pinode-without-noise}
\end{figure}

Without noise, the model reproduces the correct spike count but shows mildly attenuated oscillations as shown in Figure~\ref{fig:pinode-without-noise}. The gating variables flatten, particularly the \textit{m}-gate troughs whose subthreshold curvature becomes less sharply resolved, yielding higher RMSE values (RMSE(n, m) = \num{0.03}, \num{0.16}). Interestingly, h(t) exhibits a slightly improved RMSE of (\num{0.04}, suggesting that smoother dynamics may aid the recovery of slower inactivation processes.

\vspace{-2em}

Although the inclusion of deterministic noise slightly smooths the fast gating dynamics, it seems to act as an inductive bias toward globally stable, amplitude-preserving trajectories by regularising stiffness and suppressing overfitting to high-frequency residuals.

In contrast, the noise-free regime improves local morphological fidelity, recovering the fine subthreshold curvature of the m-gate’s activation manifold that is typically lost in low-order solvers.
This observation indicates that the end-to-end learning of the PI-NODE-SR vector field allows the network to implicitly correct solver diffusion errors, enabling a second-order Heun integrator to recover morphological features usually reserved for higher-order methods.

Together, these mechanisms make the Heun-trained PI-NODE-SR \textbf{behave as if it had a higher effective local order on the fast activation manifold}, while structured noise provides a complementary inductive bias that improves long-term temporal precision.

\subsection{Vanilla PINN with and without Normalisation}

Finally, we evaluate vanilla PINNs trained on the HH equations with and without normalisation. The same network architecture, optimiser, and training duration are used in both cases (Adam Optimiser with lr=0.001).

\begin{figure}
\centering
\begin{subfigure}{0.31\textwidth}
  \includegraphics[width=\linewidth]{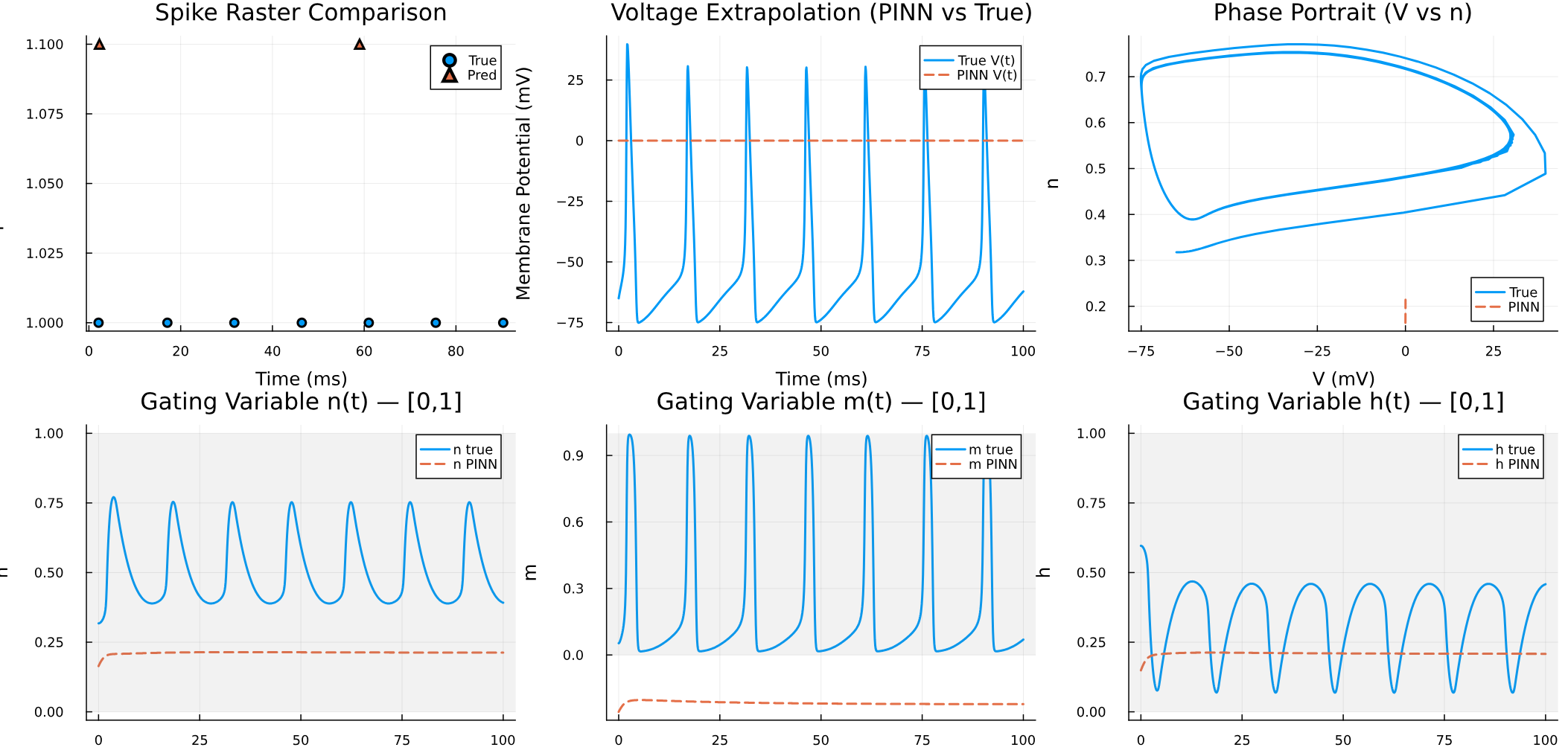}
  \caption{With normalisation}\label{fig:vanilla-pinn-with}
\end{subfigure}
\hspace{0.02\textwidth}
\begin{subfigure}{0.31\textwidth}
  \includegraphics[width=\linewidth]{figures/vanilla_PINN_results.png}
  \caption{Without normalisation}\label{fig:vanilla-pinn-without}
\end{subfigure}
\caption{\textbf{Results of training vanilla PINN with and without normalisation on HH model.} For each - Top left: spike raster comparison; Top center: predicted vs. ground truth membrane potential $V(t)$; Top right: phase portrait ($V$–$n$); Bottom: gating variable trajectories ($n$, $m$, $h$).}
\label{fig:vanilla-pinn-comparison}
\end{figure}

\begin{itemize}[itemsep=0.5\baselineskip, topsep=0pt, parsep=0pt, partopsep=0pt] \item \textbf{With normalisation:} Slightly more dynamic but voltage dynamics collapse to flat traces, producing zero spikes as shown in Figure \ref{fig:vanilla-pinn-with}. Gating variables remain static, with RMSE(m) = \num{0.43}. \item \textbf{Without normalisation:} Worse gating trajectories, RMSE(n) = \num{0.72}, and poorer phase representation. Subthreshold gating trajectories consistently observed and phase-space behavior ($V$–$n$) converges toward a degenerate point as shown in Figure \ref{fig:vanilla-pinn-without}. \end{itemize}

\vspace{2em}

Our results show that during the initial \num{5000} iterations, both normalised and unnormalised vanilla PINNs collapsed to static, non-oscillatory trajectories. Input–output normalisation alone does not significantly improve early convergence or stiffness handling under the same conditions as the PINODE ablation. Although it may aid long-term gradient stability, normalisation does not mitigate the early optimisation difficulty or the imbalance between slow and fast dynamics inherent to the Hodgkin–Huxley system.

\subsection{Summary of Findings}

Across all ablations, we find the following:
\begin{itemize}[itemsep=0.5\baselineskip, topsep=0pt, parsep=0pt, partopsep=0pt]
    \item Physics loss is necessary to prevent collapse and ensure long-term extrapolation.
    \item Adjoint choice critically affects gradient stability; interpolating adjoints enable learning through spikes, while backsolve adjoints fail.
    \item Noise regularisation improves spike timing, amplitude fidelity, and gating reconstruction.
    \item Normalisation in PINNs degrades performance by distorting physical scales.
\end{itemize}

Together, these results demonstrate that every design choice in PI-NODE-SR contributes to stable, biologically faithful extrapolation; something no baseline achieves, even when enhanced with normalisation over 5000 iterations.

\section{Discussion}

\subsection{Stability in stiff regimes}
Our experiments confirm that standard NeuralODEs and PINNs are unreliable for stiff oscillatory systems such as Hodgkin–Huxley. While the models may succeed locally, they collapse under long-horizon extrapolation, either drifting in phase or converging to fixed points, with noticeably worse training on PINN's. This aligns with existing research on the instability in NeuralODEs observed under stiffness ~\cite{NEURIPS2021_df438e52}. This instability reflects a deeper issue; residual minimisation without scale-awareness biases the optimisation toward fast variables, while slower processes vanish. Our finding that PI-NODE-SR maintains limit cycles over 7× extrapolation suggests that stiffness is as much an optimisation imbalance as it is a numerical integration challenge.

\subsection{Combining Physics-informed regularisation and explicit lower-order solvers for enhanced performance}
The results demonstrate that scale-aware residual normalisation is essential for stable physics-informed training. Interestingly, stability emerged with Heun’s method, an explicit solver widely considered unsuitable for stiff ODEs. This suggests that normalisation shifts the burden away from the solver alone, allowing lower-order methods to perform more reliably in stiff domains when paired with principled residual balancing. This challenges prevailing assumptions that only high-order implicit solvers (e.g., Radau, Rodas) are viable in these settings.

\subsection{Data efficiency and robustness}
The ability of PI-NODE-SR to extrapolate >\SI{100}{\milli\second} from a single \SI{14.68}{\milli\second} training window highlights the power of inductive biases in stiff dynamical learning. This is particularly notable under Ornstein–Uhlenbeck noise, where the model preserved spiking frequency and amplitude. Such robustness indicates that scale-aware constraints not only aid in stabilising optimisation, but also provide a safeguard against overfitting to noise - a property critical for real biological recordings.

\subsection{Limitations}
While our approach improves stability and interpretability, PI-NODE-SR remains sensitive to initialisation and does not guarantee exact recovery of biophysical parameters. Our approach also assumes full state observability, limiting direct application to experimental settings where only voltage is measured. Thus, generalising to partial observation remains an open challenge. More broadly, while residual normalisation stabilises training, it does not eliminate all sources of instability; extreme stiffness or very long extrapolations may still require implicit solvers or hybrid formulations

\section{Future Work}

This work opens several avenues for extending physics-informed neural modeling in stiff biological systems:

\paragraph{Hidden-variable discovery:}
Relaxing the assumption of fully known dynamics, future work could embed neural surrogates into unknown components of the Hodgkin–Huxley equations, such as gating kinetics, via Universal Differential Equations (UDEs). This would allow recovery of missing physics from data while preserving known structures.

\paragraph{Partial observability:}
Our current framework assumes access to all state variables $(v, n, m, h)$. Many real intracellular and electrophysiological recordings typically measure only membrane voltage. Extending PI-NODE-SR to infer latent gating variables from voltage alone through latent ODE formulations, variational inference, or physics-constrained autoencoders would substantially broaden applicability to experimental and clinical neuroscience.

\paragraph{Scalability and uncertainty:}
Biological systems exhibit variability across neurons and individuals. Incorporating stochastic differential equation (SDE) formulations and Bayesian extensions into PI-NODE-SR could enable uncertainty-aware predictions, capturing variability in spike timing, amplitude, and gating dynamics.

\paragraph{Solver and adjoint innovations:}
Our findings suggest that lower-order, explicit solvers can succeed in stiff domains when paired with scale-aware residuals. Future work should test whether this solver–loss combination generalises beyond both Heun's solver and the Hodgkin–Huxley model, potentially leveraging reversible architectures or Enzyme-based custom adjoints for scalable training.

\paragraph{Multi-compartment and network-level modeling:}
Beyond single-compartment HH neurons, our framework could be extended to simulate dendritic integration, spatially distributed neuron models, or cortical microcircuits. This would require coupling PI-NODE-SR with PDE-based or graph-based neural differential equations, offering a path toward scalable hybrid models of network-level biophysics.

\section{Conclusion}
We have demonstrated that below \num{5000} iterations of training, vanilla NeuralODEs and PINNs struggle to learn and extrapolate stiff, oscillatory biological dynamics, such as those governed by the Hodgkin–Huxley model. Our results show that stabilising Neural ODEs in stiff oscillatory domains does not require abandoning lower-order explicit solvers. Instead, pairing scale-aware residual normalisation with physics-informed training yields reliable extrapolation of long-horizon dynamics, even on challenging systems like the Hodgkin-Huxley model. Notably, we find that end-to-end learning of the vector field allows the network to implicitly correct solver diffusion errors, enabling a second-order Heun integrator to recover morphological features, such as fast gating curvature, typically reserved for higher-order methods.
Through a comprehensive evaluation and ablation study, we show that each architectural choice contributes to model stability and fidelity, suggesting that hybrid models combining data and domain knowledge offer a principled path toward robust, interpretable neural modeling of biological systems. More broadly, our findings suggest that stiffness in Neural ODE learning arises not only from integration, but also from imbalanced optimisation landscapes and that balancing physical and numerical scales provides a general recipe for both robustness and biological realism.

\bibliographystyle{plainnat}      % alphabetical order with natbib
\bibliography{references}         %
% \bibliographystyle{unsrtnat}
% \bibliography{references}  %%% Uncomment this line and comment out the ``thebibliography'' section below to use the external .bib file (using bibtex) .





% \end{thebibliography}

\end{document}